\documentclass[11pt,a4paper]{article}
\usepackage[draft]{hyperref}
\usepackage{acl2017}
\usepackage{times}
\usepackage{url}
\usepackage{latexsym}
\usepackage{microtype}
\usepackage[small]{caption}
\usepackage{amsmath}
\usepackage{amsfonts}
\usepackage{graphicx}
\usepackage{adjustbox}
\usepackage{booktabs}
\usepackage{subcaption}
\usepackage{multirow}
\usepackage{topcapt}
\usepackage[utf8]{inputenc}

\usepackage[disable]{todonotes}   
   
\newcommand{\UNK}{\mbox{$<$\hspace{-0.25ex}unk\hspace{-0.25ex}$>$}}
\newcommand{\BPE}{\mbox{\hspace{-0.25ex}@\hspace{-0.5ex}@}}

\usepackage{cleveref}
\crefname{section}{\S}{\S\S}
\Crefname{section}{\S}{\S\S}
\crefname{table}{Table}{}
\crefname{figure}{Figure}{}
\crefname{algorithm}{Algorithm}{}
\crefname{equation}{Equation}{}
\crefname{appendix}{Appendix}{}
\crefformat{section}{\S#2#1#3}  

\newlength\shlength
\newcommand\xshlongvec[2][0]{\setlength\shlength{#1pt}%
  \stackengine{-5.6pt}{$#2$}{\smash{$\kern\shlength%
    \stackengine{7.55pt}{$\mathchar"017E$}%
      {\rule{\widthof{$#2$}}{.57pt}\kern.4pt}{O}{r}{F}{F}{L}\kern-\shlength$}}%
              {O}{c}{F}{T}{S}}

\aclfinalcopy

\title{How Robust Are Character-Based Word Embeddings in Tagging and MT\\Against Wrod Scramlbing or Randdm Nouse?}

\author{Georg Heigold \\
DFKI \& Saarland University \\
Saarbr\"ucken, Germany \\
  {\tt georg.heigold@dfki.de} \\\And
  G\"unter Neumann \\
  DFKI \\
Saarbr\"ucken, Germany \\
  {\tt neumann@dfki.de} \\\And
Josef van Genabith \\
  DFKI \& Saarland University \\
Saarbr\"ucken, Germany \\
  {\tt josef.van\_genabith@dfki.de} \\}

\begin{document}
\maketitle
\begin{abstract}
This paper investigates the robustness of NLP
against perturbed word forms. 
While neural approaches can achieve (almost) human-like accuracy for certain tasks and conditions,
they often are sensitive to small changes in the input such as non-canonical input (e.g., typos).
Yet both stability and robustness are desired properties in applications
involving user-generated content,
and the more as humans easily cope with such noisy or adversary conditions.
In this paper, we study the impact of noisy input.
We consider different noise distributions
(one type of noise, combination of noise types) and 
mismatched noise distributions for training and testing.
Moreover, we empirically evaluate the robustness of 
different models (convolutional neural networks, recurrent neural networks,
non-neural models), 
different basic units (characters, byte pair encoding units), and
different NLP tasks (morphological tagging, machine translation).
\end{abstract}

\section{Introduction}

In this paper, we study the effect of non-normalized text on natural language processing (NLP).
Non-normalized text includes non-canonical word forms, 
noisy word forms, and word forms with "small" perturbations,
such as informal spellings, typos, scrambled words.
Compared to normalized text, the variability of non-normalized text
is much greater and aggravates the problem of data sparsity.

Non-normalized text dominates in many real world applications.
Similar to humans, ideally NLP should perform reliably and robustly also under
suboptimal or even adversarial conditions, without a significant
degradation in performance.
Web-based content and social media are a rich source for
noisy and informal text.
Noise can also be introduced in a downstream NLP application
where errors are propagated from one module to the next.
For example, speech translation where the machine translation (MT)
module needs to be robust against errors introduced by the
automatic speech recognition (ASR) module.
Moreover, NLP should not be vulnerable to adversarial examples.
While all these examples do not pose a real challenge to an experienced human reader, 
even "small" perturbations from the canonical form can make a state-of-the-art NLP system fail.

To illustrate the typical behavior of state-of-the-art NLP 
on normalized and non-normalized text, 
we discuss an example in the context of neural MT (NMT). 
Different research groups have shown that NMT can generate natural and fluent
translations \cite{DBLP:journals/corr/BentivogliBCF16}, 
achieving human-like performance in certain settings \cite{DBLP:journals/corr/WuSCLNMKCGMKSJL16}.
The state-of-the-art NMT engine \emph{Google Translate}\footnote{https://translate.google.com/, February 2017}, 
for example, perfectly translates the English sentence

\begin{small}\noindent 
I used my card to purchase a meal 
on the menu and the total on my 
receipt was \$ 8.95 but when I went 
on line to check my transaction it 
show \$ 10.74 .
\end{small}

\noindent
into the German sentence

\begin{small}\noindent
Ich benutzte meine Karte , um eine 
Mahlzeit auf der Speisekarte zu
kaufen und die Gesamtsumme auf
meiner Quittung war \$ 8,95 , aber 
als ich online ging , um meine 
Transaktion zu überprüfen , zeigt 
es \$ 10,74 .
\end{small}

\noindent
Adding some noise to the source sentence by swapping
a few neighboring characters, e.g.,

\begin{small}\noindent
I used my card ot purchase a meal
no the mneu and the total no my 
receipt was \$ 8.95 but whne I went
on line to check ym transaction it
show \$ 1.074 .
\end{small}

\noindent
confuses the same NMT engine considerably:

\begin{small}\noindent
Ich benutzte meine Karte ot Kauf 
eine Mahlzeit nicht die Mneu und
die insgesamt nicht meine Quittung 
war \$ 8,95 aber whne ging ich auf 
Linie zu überprüfen ym Transaktion 
es \$ 1.074 .
\end{small}

\noindent
By contrast, an experienced human reader can still
understand and correctly translate the noisy sentence
and compensate for some information loss 
(including real word errors such as "no" vs. "on", but rather not "10.74" vs. "1.074"),
with little additional effort and often not even noticing "small" perturbations.


One might argue that a good translation should in fact
translate corrupted language into corrupted language.
Here, we rather adopt the position that the objective 
is to preserve the intended content and meaning of a sentence regardless of noise.

It should be noted that neural networks with sufficient capacity,
in particular recurrent neural networks, are universal function approximators \cite{Schafer:2006:RNN:2125268.2125345}.
Hence, the performance degradation on non-normalized text is not so much
a question whether the model can capture the variability but rather how to train a robust model.
In particular, it can be expected that
training on noisy data will make NLP more robust, 
as it was successfully demonstrated for other application domains including
vision \cite{DBLP:journals/taslp/CuiGK15} and 
speech recognition \cite{multistyle-google:2016}.

In this paper, we empirically evaluate the robustness of 
different models (convolutional neural networks, recurrent neural networks,
non-neural models), 
different basic units (characters, byte pair encoding units), and
different NLP tasks (morphological tagging, NMT).
Due to easy availability and to have more control on the experimental setup
with respect to error type and error density,
we use synthetic data generated from existing clean corpora by perturbing
the word forms. The perturbations include character flips and swaps of
neighboring characters to imitate typos, and word scrambling.

The contributions of this paper are the following.
Our experiments confirm that (i) noisy input substantially degrades the output of models trained on clean data. 
The experiments show 
that (ii) training on noisy data can help models achieve performance on noisy data similar to that of models trained on clean data tested on clean data, 
that (iii) models trained noisy data can achieve good results on noisy data almost without performance loss on clean data, 
that (iv) \emph{error type} mismatches between training and test data can have a greater impact than \emph{error density} mismatches, 
that (v) character based approaches are almost always better than byte pair encoding (BPE) approaches with noisy data, 
that (vi) the choice of neural models (recurrent, convolutional) is not as significant, and 
that (vii) for morphological tagging, under the same data conditions, the neural models outperform a conditional
random field (CRF) based model.

The remainder of the paper is organized as follows.
Section \ref{sec:related} discusses related work.
Section \ref{sec:noise} describes the noise type 
and Section \ref{sec:model} briefly summarizes the modeling approaches
used in this paper.
Experimental results are shown and discussed in Section \ref{sec:experiments}.
The paper is concluded in Section \ref{sec:conclusion}.

\section{Related Work}\label{sec:related}	
A large body of work on regularization techniques to learn robust representations and models exists.
Examples include $\ell_2$-regularization, dropout \cite{DBLP:journals/corr/abs-1207-0580}, 
Jacobian-based sensitivity penalty \cite{Dauphin-et-al-NIPS2011,DBLP:journals/corr/LiCB16a},
and data noising. 
Compared to other application domains such as vision \cite{lecun-98,DBLP:journals/corr/GoodfellowSS14} and 
speech \cite{lippman1987,tuske2014data,DBLP:journals/taslp/CuiGK15,multistyle-google:2016},
working on noisy data \cite{Gimpel:2011:PTT:2002736.2002747,Derczynski:2013,DBLP:journals/corr/Plank16} 
and in particular data noising \cite{Yitong2017},
 do not have a long and extensive history in NLP.

While invariance transformations such as rotation, translation in vision
or vocal tract length, reverberation, and noise in speech have all been harnessed,
we do not have a good intuition on useful perturbations for written language yet.
Label dropout and flip (cf. typos) have been proposed both on 
the byte-level \cite{DBLP:journals/corr/GillickBVS15}  and
the word-level \cite{1703.02573}.
Syntactic and semantic noise for semantic analysis was studied in \cite{Yitong2017}.
From a human perception perspective,
word scrambling may be of interest 
\cite{Rawlinson:1976,Rayner:2006}.


The arbitrary relationship between the orthography of a word and its 
meaning in general is a well known assumption in linguistics \cite{Saussure:1916}.
However, the word form often carries additional important information.
This is, for example, the case in morphologically rich languages or in non-normalized text
where small perturbations result in similar word forms.
Recently, sub-word units have attracted some attention in NLP to handle rarely and unseen words
and to reduce the computational complexity in neural network approaches
\cite{ling-EtAl:2015:EMNLP2,DBLP:journals/corr/GillickBVS15,DBLP:journals/corr/SennrichHB15,DBLP:journals/corr/ChungCB16,heigold2017}.
Examples for sub-word units include BPE based units \cite{DBLP:journals/corr/SennrichHB15},
characters \cite{ling-EtAl:2015:EMNLP2,DBLP:journals/corr/ChungCB16,heigold2017} or
even bytes \cite{DBLP:journals/corr/GillickBVS15}.
A comparison of BPE and characters for machine translation regarding grammaticality 
can be found in \cite{DBLP:journals/corr/Sennrich16}.

\section{Noise Types}\label{sec:noise}
In this work, we experiment with three different noise types:
character swaps, character flips, and word scrambling.
Character flips and swaps are rough approximations to typos.
Word scrambling is motivated from psycholinguistic studies \cite{Rawlinson:1976}.
This choice of noise types allows us to automatically generate noisy text with different type and density distributions from existing properly edited "clean" corpora.
Using synthetic data is clearly suboptimal, 
but we use synthetic data because of their easy availability and
because it gives us better control on the experimental setup.

\paragraph*{Character swaps}
This type of perturbation randomly swaps two neighboring characters in a word.
The words are processed from left to right.
A swap is performed at each position with a pre-defined probability.
Hence, movements from the left to the right beyond neighboring characters are possible.
A character-swapped version (10\% swapping probability)
of the clean example sentence in the introduction may look like this:

\begin{small}\noindent
I used my card ot purchase a meal
no the mneu and the total no my 
receipt was \$ 8.95 but whne I went
on line to check ym transaction it
show \$ 1.074 .
\end{small}

\paragraph*{Word scrambling}
Humans appear to be good at reading scrambled text\footnote{
\url{http://www.mrc-cbu.cam.ac.uk/people/matt-davis/cmabridge/}, note the word scramble in the URL!}.
In a word scramble, the characters can be in an arbitrary order.
The only constraint is that the first and last character be at the right place.
In particular, all word scrambles are assumed to be equally likely.
A scrambled version of the clean example sentence in the introduction may look like this:

\begin{small}\noindent
I uesd my card to pchasure a mael 
on the mneu and the ttaol on my 
repciet was \$ 89.5 but wehn I went 
on line to chcek my tanrsactoin it 
sohw \$ 1.074 .
\end{small}

\noindent
Clearly, some word scrambles are easier than others.
Word scrambling approximately includes character swaps.

\paragraph*{Character flips}
This type of perturbation randomly replaces a character with another character at a pre-specified rate.
Characters are drawn uniformly, but special symbols (e.g., end of stream) are excluded. 
We do not assume any correlation across characters.
A character-flipped version (10\% flipping probability)
of the clean example sentence in the introduction may look like this:

\begin{small}\noindent
I used my car> to purch.s' a meal
on the menu and the total on my
receipv tas \$ 8.95 but whe3 = wen+
on lin4 to chece my tran\&awtion it
shzw \$ 10.74 .
\end{small}

Character flips preserve the order of characters but replace some information with random information,
whereas character swaps and word scrambling relax the order of characters but do not add
random information.

Other simple perturbations include randomly removing or adding (in particular, repeating) characters.

In the experimental section, we will consider different noise distributions
(one type of noise, combination of noise types) and 
mismatched noise distributions for training and testing.

A word of length $n$ with at most one character flip can have up to $nC$ different word forms,
where $C$ denotes the number of characters in the vocabulary.
Word scrambling multiplies the number of word forms by a factor of $(n-2)!$.
In general, perturbing word forms introduces a great deal of variability and data becomes much more sparse,
implying that efficient handling of rare and unseen words will be crucial.

\section{Modeling}\label{sec:model}

This section briefly summarizes the modeling approaches used in this work.

First, we address the choice of unit.
As illustrated in Table~\ref{tab:example} on an example from the UD English corpus\footnote{\url{http://dependencies.org/}}, 
a word-based unit does not seem to be an appropriate unit in the presence of perturbations. 
Any change of the word form implies a different, independent word index.
Even worse, most perturbed word forms do not represent valid words and 
are mapped to the \UNK-token and no word-specific information is preserved.
This suggests the use of sub-word units.
Here, we use BPE units \cite{DBLP:journals/corr/SennrichHB15} and characters as the basic units.
BPE is based on character co-occurrence frequency distributions and has the effect of representing
frequent words as whole words and splitting rare words into sub-word units
(e.g., "used" as "used", "purchase" as "purcha\BPE se").
BPE provides a good tradeoff between modeling efficiency 
(i.e., the model does not need to learn for the frequent words how to assemble them) 
and handling unknown words.
However, BPE may not be efficient at representing noisy word forms as
small perturbations can lead to a different  representation using different BPE units
(e.g., "used" vs. "u\BPE es\BPE d", "purcha\BPE se" vs. "p\BPE cha\BPE sure").
As the example illustrates (Table~\ref{tab:example}), perturbations tend to break
longer units into smaller units, which makes the use of whole word units less useful.
Finally, characters as the basic units have similar representations for similar word forms,
but result in longer sequences, which makes the modeling of long-range dependencies
harder and increases the computational complexity.
\begin{table*}[htbp]
\small
\caption{Clean (left) vs. scrambled (right) example sentence using a word-based (top), a BPE-based (middle), and a character-based (bottom) representation}
\begin{minipage}{\textwidth}
\begin{minipage}{0.49\textwidth}
I used my card to purchase a meal 
on the menu and the total on my 
receipt was \$ 8.95 but when i went 
on line to check my transaction it 
show \$ 10.74 .
\end{minipage}
\hspace{0.02\textwidth}
\begin{minipage}{0.49\textwidth}
I \UNK{}  my card to \UNK{}  a \UNK{}  
on the \UNK{}  and the \UNK{}  on my 
\UNK{}  was \$ 89.5 but \UNK{}  i went 
on line to \UNK{}  my \UNK{}  it 
\UNK{}  \$ 1.074 .
\end{minipage}
\vspace{1ex}
\end{minipage}
\begin{minipage}{\textwidth}
\begin{minipage}{0.49\textwidth}
I used my c\BPE{} ard to purcha\BPE{} se a me\BPE{} al on the men\BPE{} u and the to\BPE{} tal on my recei\BPE{} pt was \$ 8\BPE{} .\BPE{} 9\BPE{} 5 but when I went on line to check my trans\BPE{} action it show \$ 10\BPE{} .\BPE{} 7\BPE{} 4 .
\end{minipage}
\hspace{0.02\textwidth}
\begin{minipage}{0.49\textwidth}
I u\BPE{} es\BPE{} d my c\BPE{} ard to p\BPE{} cha\BPE{} sure a ma\BPE{} el on the m\BPE{} ne\BPE{} u and the t\BPE{} ta\BPE{} ol on my rep\BPE{} ci\BPE{} et was \$ 8\BPE{} 9\BPE{} .\BPE{} 5 but we\BPE{} h\BPE{} n I went on line to ch\BPE{} c\BPE
 e\BPE{} k my t\BPE{} on\BPE{} tri\BPE{} as\BPE{} ac\BPE{} n it so\BPE{} h\BPE{} w \$ 1\BPE{} .\BPE{} 0\BPE{} 7\BPE{} 4 .
\end{minipage}
\end{minipage}
\begin{minipage}{\textwidth}
\begin{minipage}{0.49\textwidth}
\begin{verbatim}
I used my card to purchase a meal on the 
menu and the total on my receipt was $ 
8.95 but when i went on line to check my 
transaction it show $ 10.74 .
\end{verbatim}
\end{minipage}
\hspace{0.02\textwidth}
\begin{minipage}{0.49\textwidth}
\begin{verbatim}
I uesd my card to pchasure a mael on the 
mneu and the ttaol on my repciet was $ 
89.5 but wehn I went on line to chcek my 
tanrsactoin it sohw $ 1.074 .
\end{verbatim}
\end{minipage}
\vspace{1ex}
\end{minipage}
\label{tab:example}
\end{table*}

Noise modeling for a word-level system is straightforward as perturbed word forms are mapped to \UNK,
i.e., noise modeling reduces to word-level label dropout (and rarely word-level label flips) \cite{1703.02573}.
This is not true for sub-word level representations, for which more detailed noise modeling will be important.

%
%
%

We use model architectures based on recurrent and convolutional neural networks in this work.
Assuming that a word segmentation is given, we first map the sub-word units of a word
to a word vector and then continue as for word-based approaches.
Deep neural networks are universal function approximators \cite{Schafer:2006:RNN:2125268.2125345}.
Hence, a neural network with sufficient capacity is expected to learn the variability induced by perturbations.
We compare the neural networks with a conditional random field \cite{Lafferty:2001:CRF:645530.655813}.

\section{Experiments}\label{sec:experiments}
In this section, we empirically evaluate the robustness against perturbed word forms 
(Section~\ref{sec:noise}) for the two common NLP tasks morphological tagging and machine translation.

\subsection{Morphological Tagging}
\begin{figure*}[htbp]
  \includegraphics[height=0.36\textwidth]{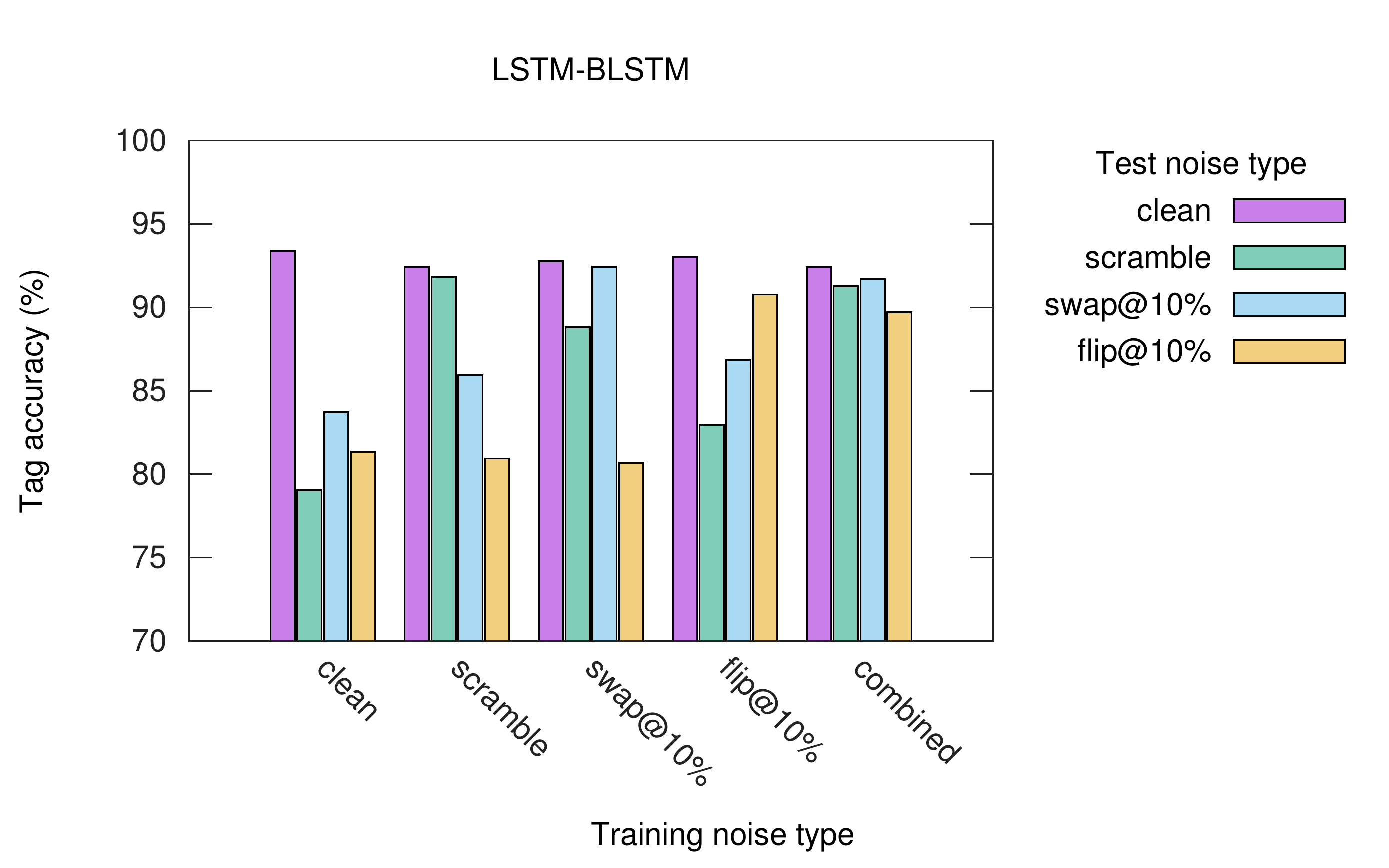}
  \includegraphics[height=0.36\textwidth]{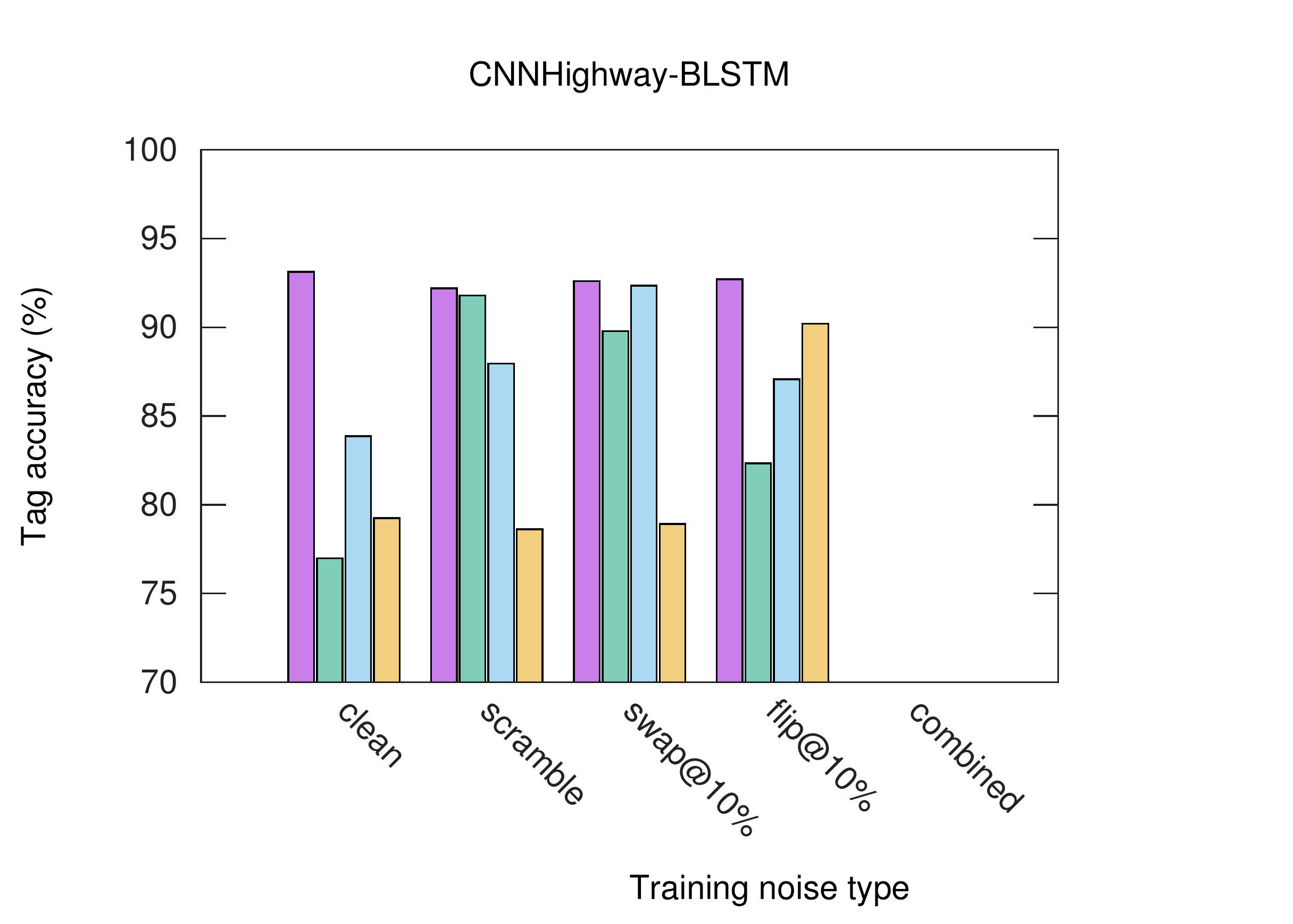}
  \includegraphics[height=0.36\textwidth]{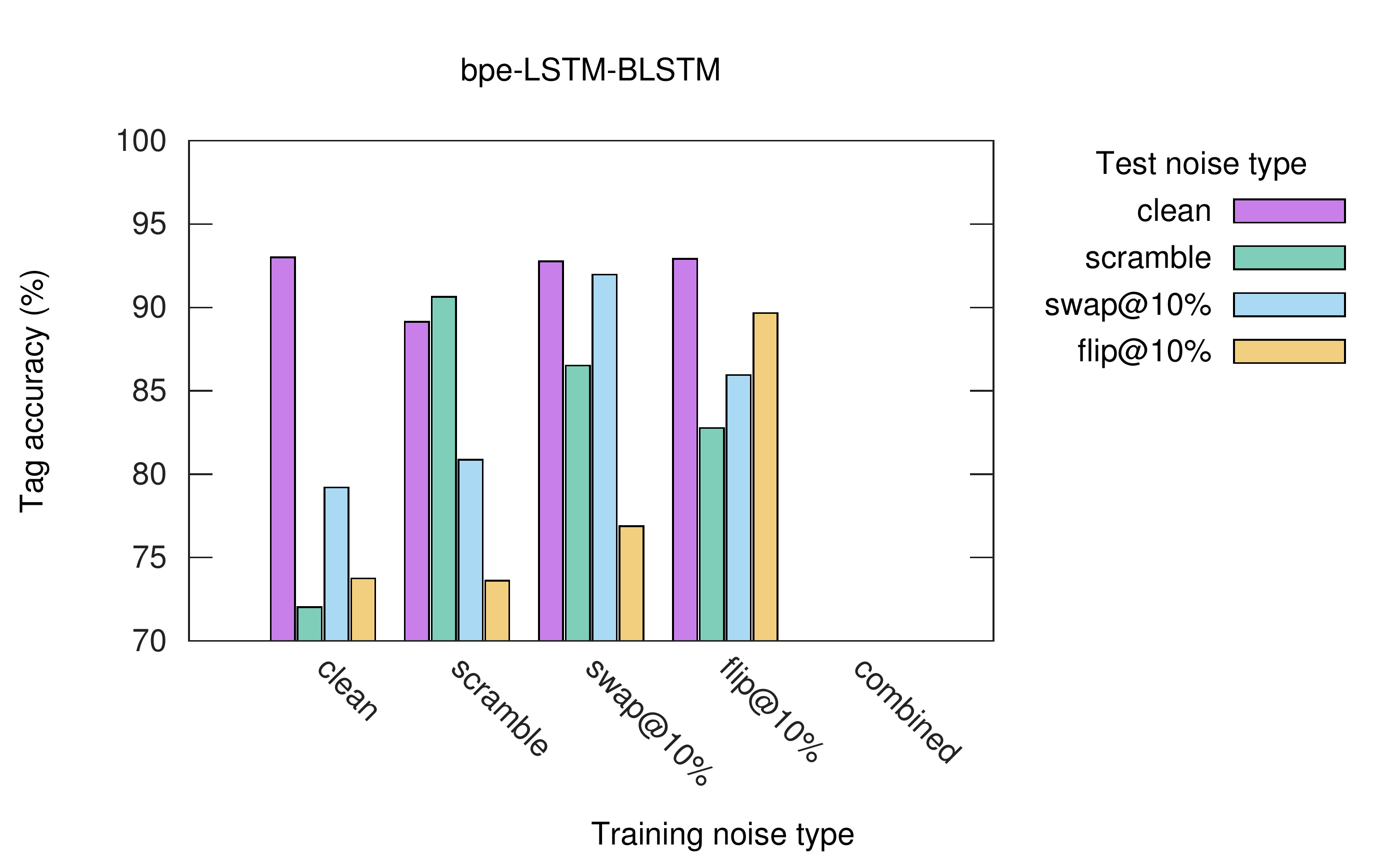}
  \includegraphics[height=0.36\textwidth]{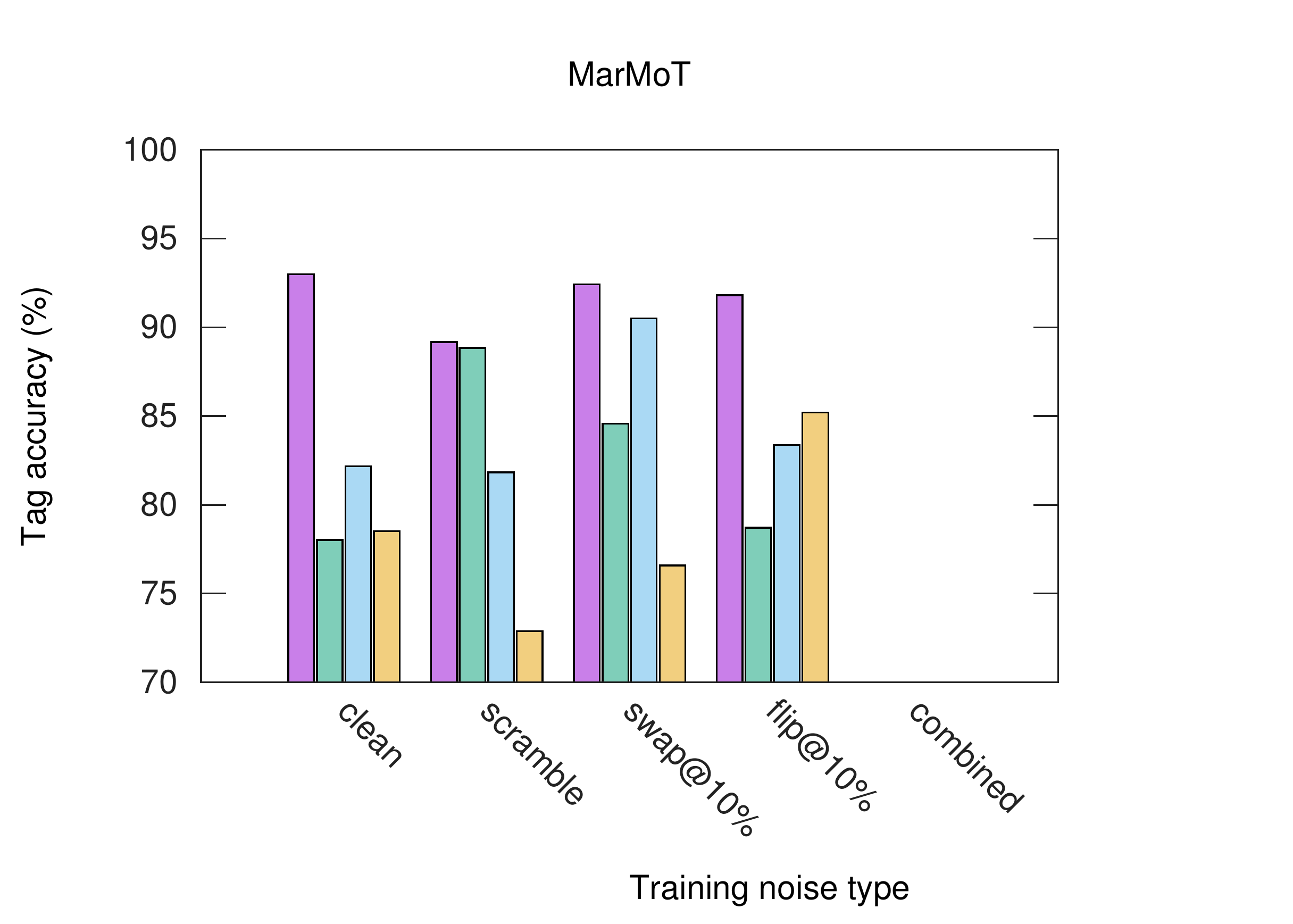}
  \caption{Noise behavior for morphological tagging for different models and units on UD English test data.
  Upper left: character-based LSTM-BLSTM.
  Upper right: character-based CNNHighway-BLSTM.
  Lower left: BPE-based LSTM-BLSTM.
  Lower right: MarMoT (CRF).}
  \label{fig:udenglish}
\end{figure*}

We used the model configurations and setups from \cite{heigold2017} 
for the morphological tagging experiments in this paper.
Training and testing was performed on the UD English data set\footnote{\url{http://dependencies.org/}}.
Figure~\ref{fig:udenglish} summarizes the results.
We explored the three main dimensions of noise type and distribution, choice of unit, and type of model.
Noise-adaptive training means standard training on noisy input sentences 
(but with correct labels: rich morphological tags or target language translation).
We distinguish the noise type and distribution used for training ("training noise type") 
and testing ("test noise type").

We start our discussion with the upper left histogram in Figure~\ref{fig:udenglish} 
for the character-based LSTM-BLSTM architecture. 
It shows a clear performance degradation from around 95\% to around 80\% tag accuracy across
all noise types compared to when trained on clean data ("clean").
Here, we consider the noise types word scrambling ("scramble", note that all words are scrambled), 
character swaps with probability 10\% ("swap@10\%),
and character flips with probability 10\% ("flip@10\%").
Bar groups 2, 3 and 4 in the upper left histogram in Figure~\ref{fig:udenglish} show that 
noise-adaptive training helps in all cases, bringing the tag accuracy back to above 90\% and without greatly affecting
the accuracy on clean data.
As expected, the accuracy under matched training and test conditions is highest in all cases.
The transferability from a noise type to another depends on the noise types.
For example, noise-adaptive training for "swap@10\%" improves the accuracy on the "scramble" test condition
by approximately 10\%. 
On the other hand, the "flip@10\%" test condition gets slightly worse.
This outcome may be expected because characters swaps are more closely related with word scrambling
than character flips.
The transferability does not need to be symmetric.
An example is "flip@10\%"-adaptive training which improves on the "swap@10\%" and "scramble" test conditions,
whereas we observe slight degradation in the opposite direction.
Finally, can we train a model that performs well across all these noise types as well as on clean data?
For this, we mixed different noise types at the sentence level for training ("combined"),
i.e., a clean sentence, followed by a sentence with scrambling inside words, 
followed by a sentence with swapped characters inside words,
followed by a sentence with flipped characters inside words, and so forth in the training data. 
The test data, by contrast, was pure clean ("clean"), scrambled ("scramble"), swapped ("swap@10\%"), or flipped ("flip@10\%") data.
According to the results summarized in the final group of bars in the upper left histogram in Figure\ref{fig:udenglish}, 
this is approximately possible.
This result again suggests that noise strongly impacts on models trained on clean data (curve for 0\% character flips), 
and that injecting noise at training time is critical but the exact noise distribution
is not so important in this case.

The upper left and lower left histograms in Figure~\ref{fig:udenglish} differ in the choice of unit on the input text side,
"char-LSTM-BLSTM" uses characters and "bpe-LSTM-BLSTM" 2,000 BPE units\footnote{
In neural MT, BPE size is usually around 50,000. 
For morphological tagging we adjust the number of BPE units according to the amount of data:
the UD English training data roughly includes 2,000 unique words with at least 10 occurrences. 
For our NMT based experiments in Section~\ref{sec:mt}, we use the customary BPE setting in NMT.}.
The overall behavior is similar, but characters seem to degrade more gracefully 
than BPE units for mismatched noise conditions
(compare bar columns 2, 3 and 4 between the upper left and lower left histograms in Figure~\ref{fig:udenglish}).

Finally, we explore how different models behave on noisy input 
(compare bar columns 2, 3 and 4 between the upper left and lower left histograms in Figure~\ref{fig:udenglish}). 
For this, we compare a char-LSTM-BLSTM, 
a char-CNNHighway-BLSTM (same as char-LSTM-BLSTM but uses a convolutional neural network
to compute the word vectors) \cite{heigold2017}, 
and a conditional random field \cite{muller-schuetze:2015:NAACL-HLT} 
including word-based features and prefix/suffix features up to length 10 for rare words
(we used MarMoT\footnote{\url{http://cistern.cis.lmu.de/marmot/}} for the experiments).
The upper left, upper right and lower right histograms in Figure\ref{fig:udenglish} show that
the qualitative behavior of the three models is very similar.
char-LSTM-BLSTM and char-CNNHighway-BLSTM achieve similar performance.
One might speculate if char-LSTM-BLSTM is slightly better at flip@10\% and char-CNNHighway-BLSTM at swap@10\% and word scrambling,
but the differences are most likely not significant.
MarMoT's tag accuracies for all noise conditions is worse by 5-10\%.

\begin{figure}
  \includegraphics[width=1.0\columnwidth]{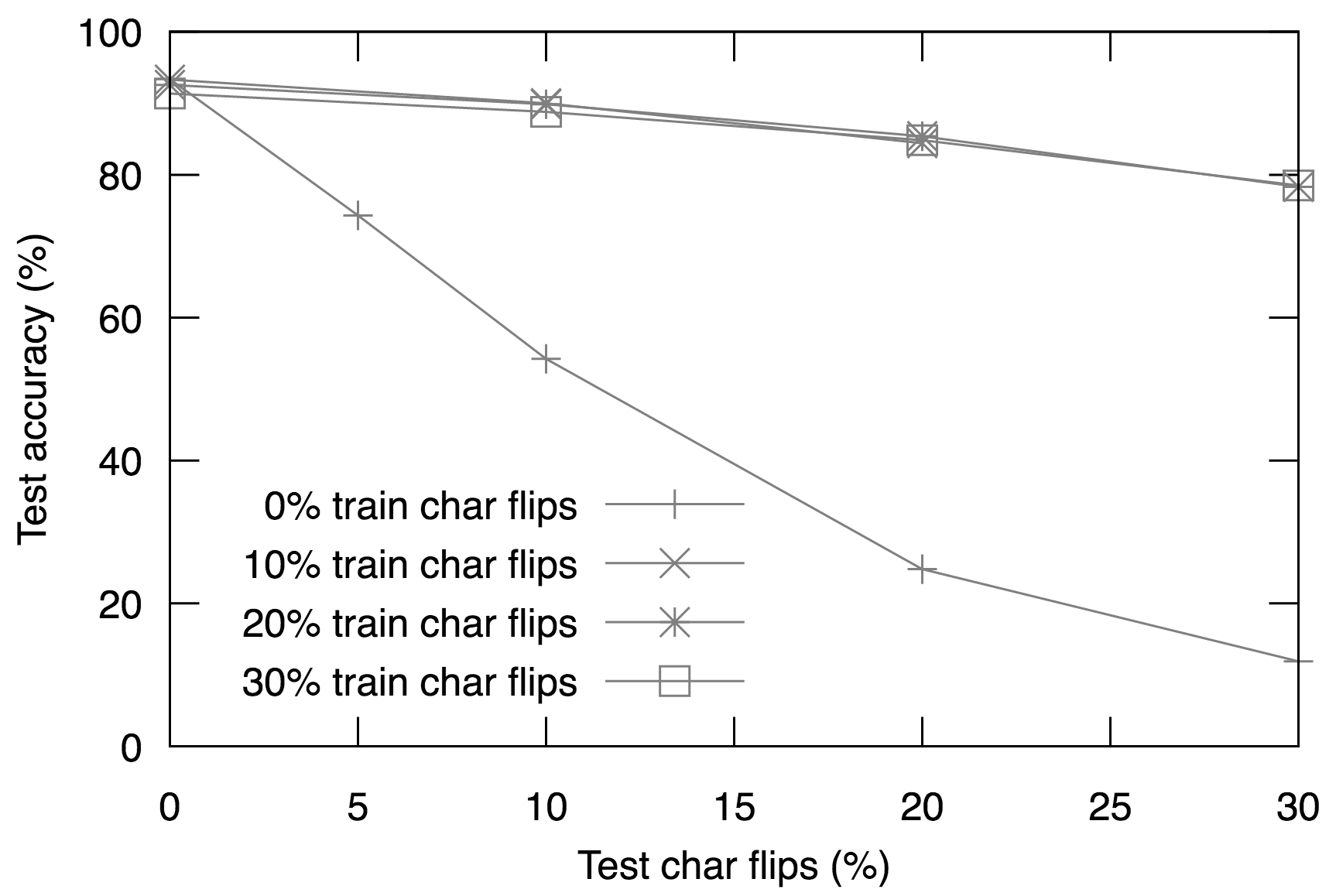}
  \caption{Effect of amount of character flips in training and testing for morphological tagging on German TIGER test data}
  \label{fig:tiger}
\end{figure}

As indicated above, Table 1 shows results on English morphological tagging. 
In a suite of experiments (not shown here in full detail for reasons of space) 
we have confirmed similar overall results for morphologically-richer languages such as German.
Morphological tagging for German is much harder than for English: 
while the English UD training data exhibit 119 distinct types of sequences of POS tags 
followed by morphological feature descriptions, 
the TIGER training data for German exhibit 681 distinct types of such sequences. 
To give one result from our German experiments, 
Figure~\ref{fig:tiger} shows the dependency of the test accuracy
on the amount of character flips in the test data, for various amounts of character flips in training.
Assuming an average word length of 6 characters, 
10\% character flips correspond with one typo in every second word,
20\% character flips with one typo per word, and
30\% character flips with two typos per word.
This result suggests that injecting noise at training time is critical,
whereas the test accuracy does not depend so much on the exact amount of training noise
(curves for 10\%, 20\% and 30\% character flips) 
and that models trained on noise injected data are still able to tag clean data with 
almost no loss in performance compared to a model trained on clean data only.

Morphological tagging is  a sequence-to-sequence labelling task, 
where (to a first approximation) the number and order of elements in the two sequences is the same 
(each word/token is paired with a POS tag plus morphological description).  
Translation is arguably a much harder task as 
it often relates sequences of different lengths with possibly substantial changes 
in the order of corresponding words/tokens between source and target and, 
compared to morphological tagging, much larger sizes of output vocabularies. 
In a second set of experiments, we explore the impact and handling of noise in the input to machine translation.

\subsection{Machine Translation}\label{sec:mt}
Our NMT setup is based on the setup in \cite{heigold2016}.
We use BPE units or characters as the basic units at the source side
and always BPE units at the target side
(following common practice in our experiments we use a BPE size of 50,000), 
resulting in the two model configurations "BPE-BPE" and "char-BPE".
For the character-based encoder, 
we assume the word segmentation and
map the word string consisting of characters or BPE units 
to a word vector by a two-layer unidirectional LSTM.
The baseline model ("clean") is trained on the German-English (DE-EN) parallel corpora
provided by WMT'16\footnote{\url{http://www.statmt.org/wmt16/translation-task.html}}.
Results for the newstest2016-deen data set are shown in Table~\ref{tab:mt-bleu}.
\begin{table}[htbp]
   \centering
   \topcaption{BLEU on newstest2016-deen for clean and noisy NMT and different test noise types} 
   \begin{tabular}{@{} lrrrr @{}} 
      \hline
      Test & \multicolumn{2}{c}{BPE-BPE} & \multicolumn{2}{c}{char-BPE}\\
      noise & & noise- & & noise- \\
      type & clean & adapted & clean & adapted \\
      \hline
      clean      & 31.6  & 30.4  & 30.7  & 30.6  \\
      swap@5\%      & 19.8  & 25.0  & 25.0  & 29.2  \\
      scramble   &  3.6  &  9.4  &  5.4  & 20.0  \\
      flip@5\%      & 16.1  & 22.5  & 21.7  & 27.1  \\
      noisy      & 21.9  & 25.6  & 21.1  & 28.5  \\
\hline
   \end{tabular}
   \label{tab:mt-bleu}
\end{table}
For noise-adaptive training, we perform transfer learning on the perturbed source sentence-target sentence pairs
("noise-adapted").
For training, we choose the following sentence-level noise distribution:
50\% clean sentences, 20\% sentences with character swaps (5\% swap probability), 
10\% sentences with word scrambles,
and 20\% sentences with character flips (5\% flip probability).
We refer to this noise distribution to as "noisy."
Beside this "noisy" noise distribution, we also use mismatched noise conditions at test time,
consisting of a single noise type only, referred to as
"clean", "swap@5\%", "scramble", and "flip@5\%".

The baseline's performance drop for noisy test data is drastic and clearly depends on the noise type.
Word scrambling seems to be the hardest noise type, for which BLEU goes down from around 30 to around 5
for BPE-BPE and char-BPE.
Overall, however, the results suggest that the char-BPE baseline degrades much more gracefully than the BPE-BPE baseline.

The results in Table~\ref{tab:mt-bleu} show that noise-adaptive training can considerably improve the performance 
on noisy data and the gap between clean and noisy conditions can be almost closed for the "easy" noise conditions.
Similar to the baseline, char-BPE tends to be less sensitive to mismatched noise conditions.
This may be best seen from the fact that char-BPE performs better or no worse than BPE-BPE for all noise conditions.
Moreover, noise-adaptive training does not affect BLEU for char-BPE (30.7 vs. 30.6) but there is a small
performance penalty for BPE-BPE (31.6 vs. 30.4).
Furthermore, the "noisy" BLEU is the highest among the noisy conditions for BPE-BPE while the "swap@5\%" BLEU is the
best for char-BPE.

We show an example for the different noise types and source representations in Table~\ref{tab:nmt-examples}.
The example reflects the general findings based on BLEU scores (Table~\ref{tab:mt-bleu}).
The example also highlights the potential difficulty of correctly translating proper names in noisy conditions.
\begin{table*}[t]
   \centering\small
   \topcaption{Example sentence for different noise types (clean, character swaps, word scrambling, character flips)
   and NMT configurations (BPE/characters and standard training/noise-adaptive training)} 
   \begin{tabular}{@{} ll @{}} 
   \hline\hline
   source & 
   Herr Modi befindet sich auf einer fünftägigen Reise nach Japan , um die wirtschaftlichen Beziehungen mit \\
   (clean) & der drittgrößten Wirtschaftsnation der Welt zu festigen . \\
\hline
unadapted & 
Mr Modi is on a five-day trip to Japan to consolidate economic
relations with the world 's third largest \\
(BPE-BPE) &  economies . \\
noise-adapted & 
Mr Modi is on a five-day trip to Japan to consolidate economic relations with the third largest economic \\
 (BPE-BPE) &  nation in the world . \\
\hline
unadapted &
Mr Modi is on a five-day trip to Japan to consolidate economic 
relations with the world 's third largest \\
(char-BPE) &  economy . \\
noise-adapted &
Mr Prodi is on a five-day trip to Japan in order to consolidate economic relations with the world 's third \\
 (char-BPE) &  largest economy. \\
\hline\hline
source & 
Herr Modi befindet sich auf einer fünftägigen Reise nach Japan, um die wirtschaftlichen Beziehungen mit \\
 (swap@5\%) & der rdtitgrößten Wirtschaftsnation der Welt zu festiegn.\\
 \hline
unadapted &
Mr Modi is on a five-day trip to Japan to entrench economic relations with the world 's most basic economic \\
 & nation .\\
noise-adapted &
Mr Modi is on a five-day trip to Japan to establish economic relations
with the world 's largest economic \\
 (BPE-BPE) & nation . \\
 \hline
unadapted &
Mr Modi is on a five-day trip to Japan to establish economic relations
with the world's largest economy. \\
 & \\
noise-adapted &
Mr Prodi is on a five-day trip to Japan in order to consolidate economic relations with the world's third \\
 (char-BPE) &  largest economy. \\
\hline\hline
source &
Hrer Modi bfdneeit scih auf eienr fggnefüiätn Reise ncah Jpaan , um die wctathhilsfecirn Buzegehnein mit \\
(scramble) & der drtiößettrgn Wsfactohtairsntin
der Welt zu fgteesin . \\
\hline
unadapted &
Hrer modes Bfdneeit scih on eienr fggnefün journey ncah Jpaan to get
the wctathsusfecirn Buzehno with the \\
 & drone Wsfactohtairsntin in the world . \\
noise-adapted &
Mr Modi is looking forward to a successful trip to Jpaan in order to
find the scientific evidence with the  \\
(BPE-BPE) & world 's largest economy in the world .\\
\hline
unadapted & 
Hear is a member of the United States of America and the United States
of America . \\
 & \\
noise-adapted &
Mr Prodi is working on a fictitious journey to Japan in order to
address economic relations with the world 's \\
(char-BPE) & third largest economy . \\
\hline\hline
source &
Herr Modi befindet sicC 0uf einer fünftägigen Reise nach Japan , u" die
wirtsch\_átlichen Beziehungen mi4 \\
(flip@5\%) &  dLr drittgrößten Wirtschaftsn,tion der Welt zu f?stigen .\\
\hline
unadapted &
Mr. Modi is located at sicC 0uf a five-day trip to Japan , u" the
wiring relations mi4 dLr third-largest \\
& economy of the world .  \\
noise-adapted &
Mr Modi is on a five-day trip to Japan to promote economic relations with the world 's third largest \\
(BPE-BPE) & economy . \\
\hline
unadapted &
Mr Modi is going to Japan on a five-day trip to Japan to fudge economic relations with the world's third \\
&  largest economy . \\
noise-adapted &
Mr Prodi is on a five-day trip to Japan to consolidate economic
relations with the world 's third largest \\
(char-BPE) &  economy . \\
\hline\hline
\end{tabular}
\label{tab:nmt-examples}
\end{table*}

\section{Conclusion}\label{sec:conclusion}
In this paper, we presented an empirical study on morphological tagging
and machine translation for noisy input.
Mostly as expected from other application domains such as vision and speech, 
we found that state-of-the-art systems are
very sensitive to slightly perturbed word forms that do not pose a challenge to humans and
that injecting noise at training time can improve the robustness of such systems considerably.
The best results were observed for matched training and test noise conditions
but generalization across certain noise types and noise distributions is possible.
Character-based approaches seem to degrade more gracefully compared with BPE-based approaches. 
We observe similar overall trends across tasks (morphological tagging and machine translation) and languages (English and German for morphological tagging). 
The results in this paper are promising but should be taken with a grain of salt
as we used augmented data based on a limited number of idealized perturbation types.
Future work will aim at a better comprehension of relevant and hard or even adversarial
perturbations and noise types (including noisy sentence structure) 
in language and testing on real noisy user input.
Moreover, the observation that the lower the BPE size is, the closer BPE is to character based encoding
and the higher the BPE size is, the closer BPE is to word based approaches,
will allow us to tune the system for the optimal granularity.
A reasonable assumption is that the denoising is task-independent and could be
trained independently of the actual NLP task, 
or shared across NLP tasks and jointly optimized.


\end{document}